%% file: DIViS.tex
\documentclass[10pt,twocolumn,letterpaper]{article}

\usepackage{cvpr}
\usepackage{times}
\usepackage{epsfig}
\usepackage{graphicx}
\usepackage{amsmath}
\usepackage{amssymb}
\usepackage{booktabs}
\usepackage{url}
\usepackage{mathtools}
\usepackage{color}
\usepackage{wrapfig}
\usepackage{caption}
\usepackage{array}
\usepackage{multirow}

\usepackage[pagebackref=true,breaklinks=true,letterpaper=true,colorlinks,bookmarks=false]{hyperref}
\cvprfinalcopy 

\def\cvprPaperID{****} 

\ifcvprfinal\fi
\begin{document}

\title{DIViS: Domain Invariant Visual Servoing for Collision-Free Goal Reaching }

\author{
  Fereshteh Sadeghi\\
  University of Washington
  \vspace{-.09in}
}

\teaser{
  \includegraphics[width=0.99\linewidth]{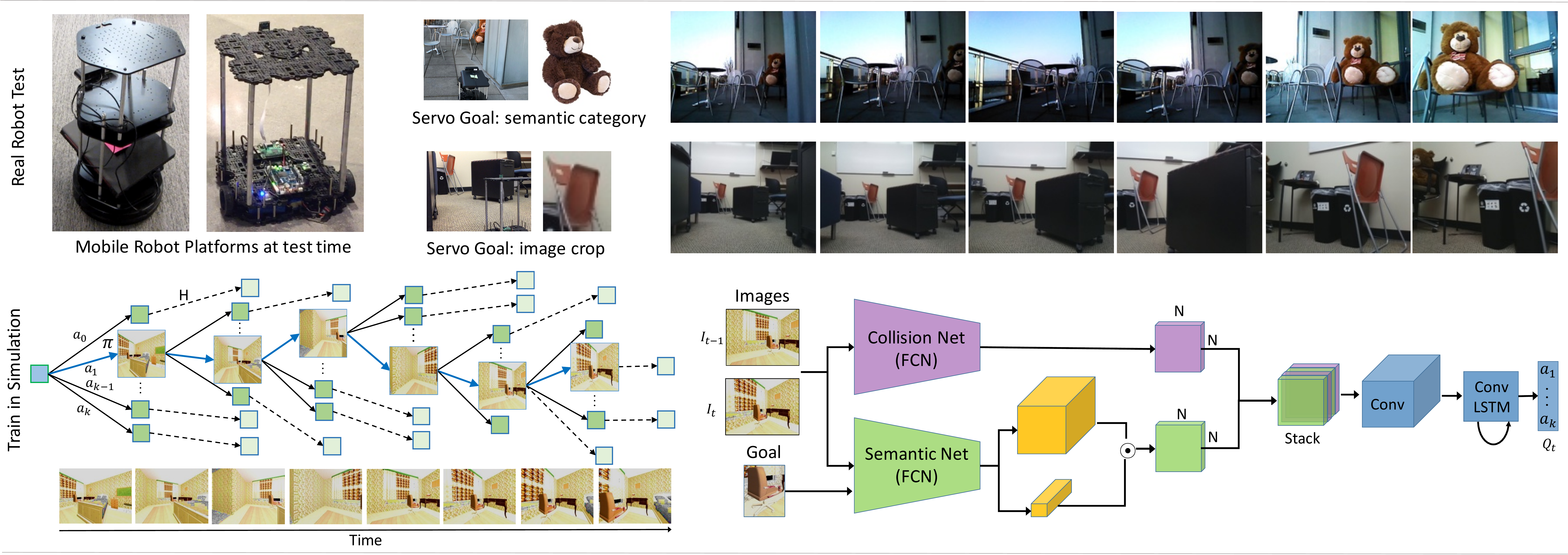}
  \vspace{-3.5 mm}
   \caption{\small{Domain Invariant Visual Servoing (DIViS) learns collision-free goal reaching entirely in simulation using dense multi-step rollouts and a recurrent fully convolutional neural network (bottom). DIViS can directly be deployed on real physical robots with RGB cameras for servoing to visually indicated goals as well as semantic object categories (top).}}
   \label{fig:net_arch}
\vspace{-2 mm}
}

\maketitle

\begin{abstract}
\vspace{-4 mm}
\input{abstract.tex}

\end{abstract}

\section{Introduction}
\input{intro.tex}

\section{Related Work}
\input{related.tex}

\section{Diverse Collision Free Goal Reaching}

\input{sim}

\section{Domain Invariant Visual Servoing}
\label{sec:approach}

\input{approach}

\section{Experimental Results}
\label{sec:result}
\input{exp.tex}


\section{Discussion}
\label{sec:conclusion}
\input{con.tex}

\vspace{0.05in}
\noindent{\bf Acknowledgment}
This work was made possible by NVIDIA support through a Graduate Research Fellowship to Fereshteh Sadeghi as well as support from Google. We would like to thank Pieter Abbeel for insightful discussions and Maya Cakmak for providing Turtlebot2 at UW for real world experiments.


{\small
\bibliographystyle{ieee}
\bibliography{DIViS}
}

\end{document}

%% file: abstract.tex
Robots should understand both semantics and physics to be functional in the real world. While robot platforms provide means for interacting with the physical world they cannot autonomously acquire object-level semantics without needing human. In this paper, we investigate how to minimize human effort and intervention to teach robots perform real world tasks that incorporate semantics. We study this question in the context of visual servoing of mobile robots and propose DIViS, a {\bf \emph{D}}omain {\bf \emph{I}}nvariant policy learning approach for collision free {\bf \emph{Vi}}sual {\bf \emph{S}}ervoing. DIViS incorporates high level semantics from previously collected static human-labeled datasets and learns collision free servoing entirely in simulation and without any real robot data. However, DIViS can directly be deployed on a real robot and is capable of servoing to the user-specified object categories while avoiding collisions in the real world. DIViS is not constrained to be queried by the final view of goal but rather is robust to servo to image goals taken from initial robot view with high occlusions without this impairing its ability to maintain a collision free path. We show the generalization capability of DIViS on real mobile robots in more than $90$ real world test scenarios with various unseen object goals in unstructured environments. DIViS is compared to prior approaches via real world experiments and rigorous tests in simulation. 
For supplementary videos, see: \href{https://fsadeghi.github.io/DIViS}{https://fsadeghi.github.io/DIViS}

\begin{figure}[t]
\centering
\includegraphics[width=1.00\linewidth]{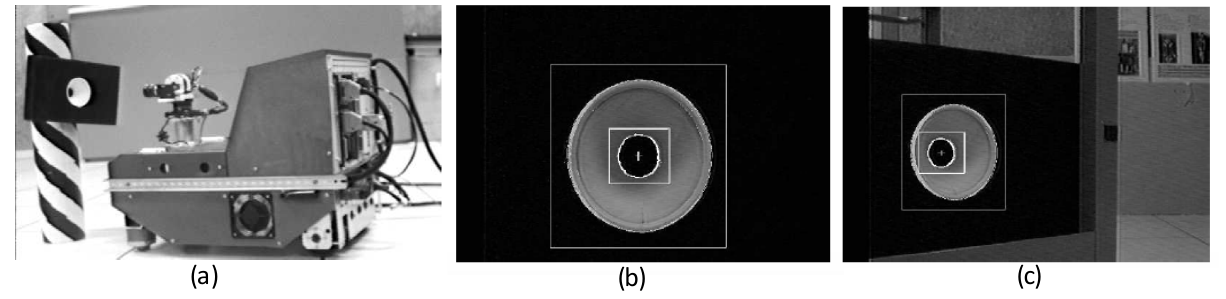}
\vspace{-0.25in}
\caption{\small{ (a) The classic 1995 visual servoing robot~\cite{pissard1995applying,espiau1992new}. The image at final position (b) was given as the goal and the robot was started from an initial view of (c).} 
   \label{fig:old_servo}
\vspace{-.5in}
}
\end{figure}

%% file: intro.tex
Perception and mobility are the two key capabilities that enable animals and human to perform complex tasks such as reaching food and escaping predators. Such scenarios require the intelligent agent to make high level 
inference about the physical affordances of the environment as well as semantics to recognize goals and distinguish walkable areas from the obstacles to take efficient actions towards the goal. Visual Servoing (VS), is a classic robotic technique that relies on visual feedback to control the motion of a robot to a desired goal which is visually specified~\cite{hutchinson1996tutorial, corke1993visual,chaumette2006visual, yoshimi1994active}. 
Historically, visual servoing has been incorporated for both robotic manipulation and navigation~\cite{hutchinson1996tutorial, corke1993visual,basri1999visual, caron2013photometric, jagersand1997experimental}. Figure~\ref{fig:old_servo} depicts an early visual servoing mobile robot in 1995 that servos to a goal image by matching geometric image features between the view at the final desired position and robot's current camera view~\cite{pissard1995applying,espiau1992new}. While visual servoing mechanisms aim to acquire the capability to surf in the 3D world, they inherently do not incorporate object semantics. In addition, conventional servoing mechanisms need to have access to the robot's view in its final goal position. Such requirement, can restrict the applicability as it may be infeasible to have the camera view at the goal position specially in unstructured and unknown real-world environments.

Deep learning has achieved impressive results on a range of supervised semantic recognition problems in vision~\cite{ksh-incdc-12,girshick2014rich} language~\cite{mikolov2013efficient} and speech~\cite{hinton2012deep,graves2013speech} where supervision typically comes from human-provided annotations. In contrast, deep reinforcement learning (RL)~\cite{mnih2015human,trpo,mnih2016asynchronous} have focused on learning from experience, which enables performance of physical tasks, but typically do not focus on widespread semantic generalization capability needed for performing tasks in unstructured and previously unseen environments. Robots, should be capable of understanding both semantics and physics in order to perform real world tasks. Robots have potential to autonomously learn how to interact with the physical world. However, they need human guidance for understanding object-level semantics (e.g., chair, teddy bear, etc.). While it is exceedingly time-consuming to ask human to provide enough semantic labels for the robot-collected data to enable semantic generalization, static computer vision datasets like MS-COCO~\cite{lin2014microsoft} or ImageNet~\cite{ksh-incdc-12} already offer this information, but in a different context. In this work, we address the question of how to combine autonomously collected robot experience with the semantic knowledge in static datasets, to produce a semantic aware robotic system. 

We study this question in the context of goal reaching robotic mobility in confined and cluttered real-world environments and introduce DIViS, {
\bf D}omain {\bf I}nvariant {\bf Vi}sual {\bf S}ervoing that can maintain a collision-free path while servoing to a goal location specified by an image from the initial robot position or a semantic object category. This is in contrast to the previous visual servoing approaches where the goal image is taken in the final goal position of the robot. Performing this task reflects robots capability in understanding both semantics and physics; The robot must be able to have sufficient physical understanding of the world to reach objects in cluttered rooms without colliding with obstacles. Also, semantic understanding is required to associate images of goal objects, which may be in a different context or setting, with objects in the test environment. In our setup, collisions terminate the episode and avoiding obstacles may necessitate turning away from the object of interest. This requires the robot to maintain memory of past movements and learn a degree of viewpoint invariance, as the goal object might appear different during traversal than it did at the beginning of the episode. This enforces the robot to acquire a kind of internal model of ``object permanence'' \cite{baillargeon1991object,bremner1994infancy,bremner2015perception}. This is similar to how infants must learn that
objects continue to exist in their previous position once
they are occluded, so too the robot must learn that an
object specified by the user continues to exist.

The main contribution of our work is a novel domain invariant policy learning approach for direct simulation to real world transfer of visual servoing skills that involve object-level semantics and 3D world physics such as collision avoidance. We decouple physics and semantics by transferring physics from simulation and leveraging human annotated real static image datasets. Therefore, our approach addresses the challenges and infeasibilities in collecting large quantities of semantically rich and diverse robot data. To keep track of changes in viewpoint and perspective, our visual servoing model maintains a short-term memory via a recurrent architecture. In addition, our method predicts potential travel directions to avoid colliding with obstacles by incorporating a simple reinforcement learning method based on multi-step Monte Carlo policy evaluation. We conduct wide range of real-world quantitative and qualitative evaluations with two different real robot platforms as well as detailed analysis in simulation. Our policy which is entirely trained in simulation can successfully servo a real physical robot to find diverse object instances from different semantic categories and in a variety of confined and highly unstructured real environments such as offices and homes.

%% file: related.tex
Visual servoing techniques rely on carefully designed visual features and may or may not require calibrated cameras~\cite{corke1993visual, espiau1992new,whb-reecu-96, ya-auvs-94,jagersand1997experimental}. Our method does not require calibrated cameras and does not rely on geometric shape cues. While photometric image-based visual servoing~\cite{caron2013photometric} aims to match a target image, our approach can servo to goals images which are under occlusion or partial view. Recently several learning based visual servoing approaches using deep RL are proposed for manipulation~\cite{levine2018learning, lampe2013acquiring, sadeghi2018sim2realservo}, navigation via a goal image ~\cite{pathak2018zero, zhu2016target} and tracking~\cite{lee2017learning}. In contrast to the the goal-conditioned navigation methods of~\cite{pathak2018zero,zhu2016target}, our approach uses the image of goal object from the initial view of the robot which can be partially viewed or heavily occluded. Also, our method learns a domain invariant policy that can be transferred to the real world despite the fact that it is entirely trained in simulation.

Transferring from simulation to real and bridging the reality gap has been an important area of research for a long time in robotics. 
Several early works include using low-dimensional state representations~\cite{mordatch2015ensemble, jakobi1995noise, cutler2015real, taylor2009transfer}. 
Given the flexibility and diversity provided by the simulation environments, learning policies in simulation and transferring to the real world has recently gained considerable interest in vision-based robotic learning~\cite{sadeghi2017cadrl,sadeghi2018sim2realservo,riedmiller2018learning,matas2018sim,james2017transferring,bousmalis2018using}. Prior works on representation learning either learn transformation from one domain to another~\cite{gopalan2011domain,rusu2016sim} or train domain invariant representations for transfer learning~\cite{long2015learning,bousmalis2016domain,ganin2016domain}. ~\cite{zhang2018vr} proposed inverse mapping from real images to renderings using CycleGAN~\cite{zhu2017unpaired} and evaluated on a self-driving application and an indoor scenario. Domain randomization in simulation for vision-based policy learning was first introduced in~\cite{sadeghi2017cadrl} for learning generalizable models that can be directly deployed on a real robot and in the real world. It was then broadly used for various robot learning tasks~\cite{sadeghi2018sim2realservo,james2017transferring,tremblay2018training,andrychowicz2018learning,pinto2017asymmetric,tobin2017domain,peng2018sim,muratore2018domain,stein2018genesis,matas2018sim}. 
In contrast to these prior works, our focus is on combining transfer from simulation (for physical understanding of the world) with transfer from semantically labeled data (for semantic understanding). The physical challenges we consider include navigation and collision avoidance, while the semantic challenges include generalization across object instances and invariance to nuisance factors such as background and viewpoint. To our knowledge, direct simulation to real world transfer for semantic object reaching with diverse goals and environments in the real world is not explored before.

Visual navigation is a closely related problem to our work. A number of prior works have explored navigation via deep reinforcement learning using recurrent policy, auxiliary tasks and mixture of multiple objectives in video game environments with symbolic targets that do not necessarily have the statistics of real indoor scenes~\cite{mnih2016asynchronous,dosovitskiy2016learning,mirowski2016learning,jaderberg2016reinforcement}. We consider a problem setup with realistic rooms populated with furniture and our goal is defined as reaching specific objects in realistic arrangements. Vision-based indoor navigation in simulation under grid-world assumption is considered in several prior works~\cite{zhu2016target, gupta2017cognitive}. ~\cite{zhang2016deep} addressed visual navigation via deep RL in real maze-shaped environment and colored cube goals without possessing semantics of realistic indoor scenes. ~\cite{savva2017minos} benchmarked basic RL strategies~\cite{dosovitskiy2016learning,mirowski2016learning,mnih2016asynchronous}. With the increasing interest in learning embodied perception for task planning and visual question answering, new simulated indoor environments have been developed~\cite{savva2017minos,wu2018building,song2017semantic,anderson2018vision,fried2018speaker}. While these environments facilitate policy learning in the context of embodied and active perception for room navigation, they do not consider generalization to the real world with a physical robot. As opposed to all aforementioned prior works, the focus in our work is on transfer of both semantic and physical information, with the aim of enabling navigation in real-world environments. In our work, we do not focus on long-term navigation (e.g., between rooms) or textual navigation~\cite{anderson2018vision}, but we focus on local challenges of highly confined spaces, collision avoidance, and searching for occluded objects which require joint understanding of physics and semantics.

\begin{figure}[t]
  \begin{center}
    \includegraphics[width=\linewidth]{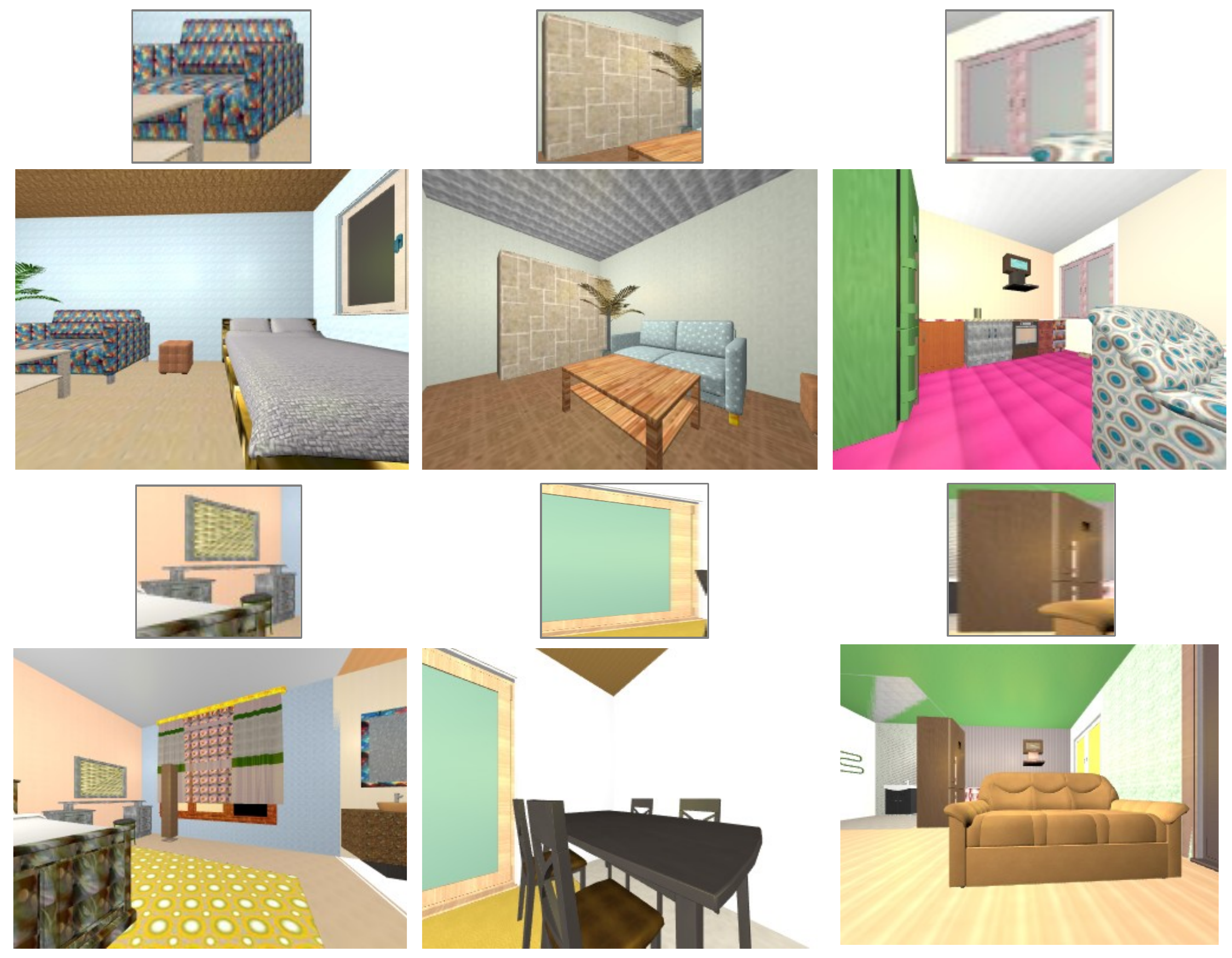}
  \end{center}
   \vspace{-.25in}
  \caption{\small{Examples of the diverse goal-reaching mobility tasks. In each task, agent should reach a different visually indicated goal object. In each scenario, we use domain randomization for rendering.}}
   \vspace{-.15in}
  \label{fig:sample_tasks}
\end{figure}

Simultaneous localization and mapping (SLAM) has been used for localization, mapping and path planning. One group of SLAM methods create a pipeline of different methods to first build a map using geometry constraints and then perform path planning for navigation in the estimated 2D or 3D map. Another group, traverse and build representation of a new environment and simultaneously plan to navigation to a goal. While these approaches provide promising solutions for building the map and navigate, their main focus is not on solving visual goal reaching and trasnfer. There is a large body of work on SLAM, for which we refer the reader to these comprehensive surveys~\cite{thrun2005probabilistic,cadena2016past}.

%% file: sim.tex
We build a new simulator for the collision-free visual goal-reaching task using a set of 3D room models with diverse layouts populated with diverse furniture placements. Our goal is to design diverse tasks that require the agent to possess both semantic and physical understanding of the world.  
We define various object-reaching navigation scenarios where in each scenario we specify a different visual goal and place the agent in a random location and with a random orientation. The agent must learn a single policy capable of generalizing to diverse setups rather than memorizing how to accomplish an specific scenario. 

Figure~\ref{fig:sample_tasks} depicts examples of our tasks. In each example, the agent is tasked to get as close as possible to a visually indicated goal while avoiding collisions with various obstacles. The map of the environment is unknown for the agent and the agent is considered to be  similar to a ground robot which can only move in the walkable areas. Collisions terminate the episode,
requiring the agent to reason about open spaces. This scenario is aligned with real-world settings where we have diverse environments populated with variable furniture. The furniture and room objects are not registered in any map, and the robot should be able to move towards objects without collision, since otherwise it can damage itself and the environment. 
Collision avoidance also enforces the agent to learn an efficient policy without needing to often backtrack.

In our setup, the robot action is composed of rotation $\tau$ and velocity $v$. We consider a fixed velocity which implies that a fix distance is traversed in each step. We discretize the rotation range into $K-1$ bins and the action at each step is the rotation degree corresponding to the chosen bin. We also add an stop action making the total number of actions equal to $K$. Note that although our action space is discrete, we move the agent in a continuous space rather than a predefined discrete grid. 
Following the success of~\cite{sadeghi2017cadrl} learning generalizable vision-based policies, we devise a randomized simulator where light, textures and other rendering setups are randomized both at training and test time. In contrast to the prior work with grid world assumption~\cite{zhu2016target,gupta2017cognitive}, our setup creates a problem with~\emph{infinite} state space which is more realistic and more challenging.

\begin{figure*}[t]
    \centering
    \includegraphics[width=\textwidth]{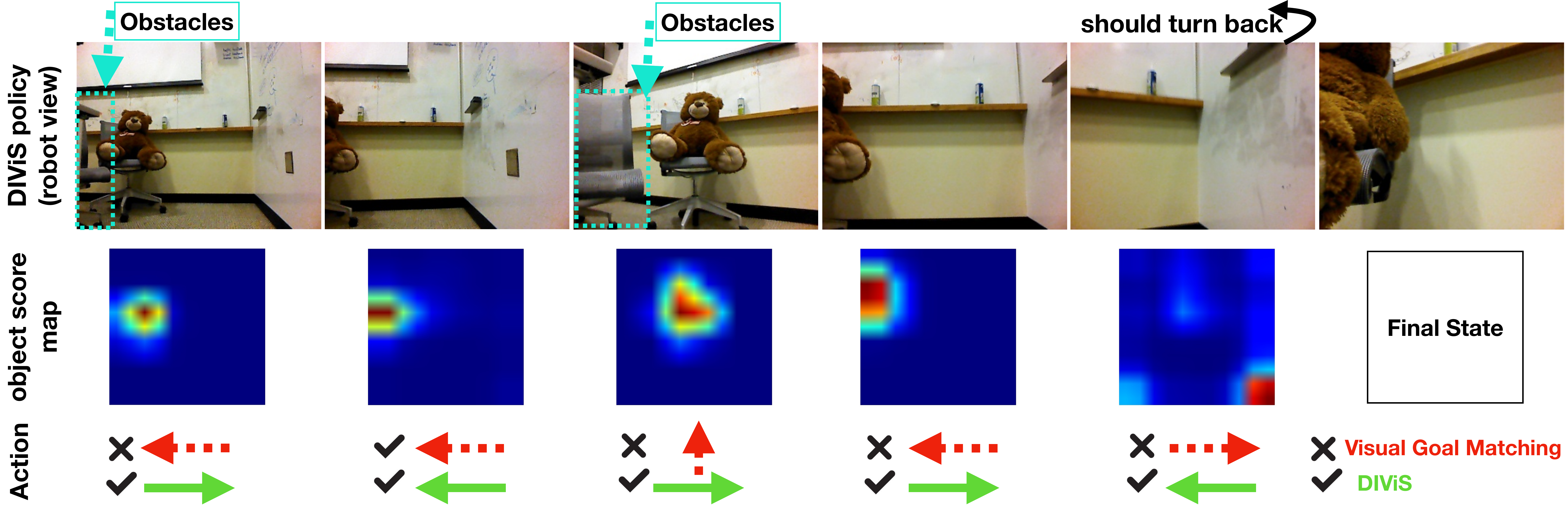}
     	\vspace{-0.25in}
    \caption{\small{Qualitative comparison of DIViS against Visual Goal Matching policy while the robot is tasked to reach ``teddy bear''. \emph{Green arrows} show action directions taken by DIViS and \emph{red arrows} (bottom row) show action directions chosen by Visual Goal Reaching which only relies on object score maps (middle row). Visual Goal Matching fails by colliding into obstacles while DIViS reaches the goal successfully by taking  turns around the obstacles (top row). }}
    \label{fig:teddy_expample}
     \vspace{-0.1in}
\end{figure*}

%% file: approach.tex
The visual servoing task of collision-free goal-reaching involves multiple underlying challenges: (1) Learning to visually localize goal objects. (2) Learning to predict collision map from RGB images. (3) Learning the optimal policy~$\pi$ to reach the objects. To integrate rich visual semantics while learning the control policy we opt to disentangle perception from control. 
We first learn a semantically rich model $\Phi_{\hat{\theta}}$ to represent visual sensory input as observation $o$.  Freezing the $\hat{\theta}$ parameters, we then learn the control policy $\pi_\theta$.

\subsection{Network Architecture}
 Our representation learning module consists of two fully convolutional networks both based on VGG16 architecture~\cite{VGG16} which we call them as \emph{Collision Net} and \emph{Semantic Net}. Figure~\ref{fig:net_arch} shows our network architecture.

For the Collision Net, we follow~\cite{sadeghi2017cadrl} to pre-train a free-space prediction network. 
Collision Net takes in an RGB image $I$ and the output logit of its last convolutional layer provides an $N\times N$ collision map $\phi_{\hat{\theta}_{c}}(I)$ which predicts if an obstacle exists in the distance of 1 meter to the agent in its 2D ego-centric view. In Figure~\ref{fig:net_arch}, $\phi_{\hat{\theta}_{c}}(I)$ is shown as a purple $N \times N$ map.

Our Semantic Net is built and trained based on~\cite{izadinia2018viser}.  
We pre-train our Semantic Net on the MS-COCO object categories~\cite{lin2014microsoft} with a weakly supervised object localization setup similar to~\cite{izadinia2018viser}. We use the penultimate layer of the fully convolutional neural network of~\cite{izadinia2018viser} to encode visual semantics in the spatial image space. Semantic Net describes each RGB image input via an $N\times N \times 2048$ map which we denote it as $S(I)$ and is shown by yellow box in Figure ~\ref{fig:net_arch}.

For each goal reaching task, our network takes in a goal image $I^{g}$ that is fixed during the entire episode. At each timestep $t$, the network takes in the current image $I_t$, the previous image $I_{t-1}$ as well as a goal image $I^{g}$. We use our Semantic Net to compute semantic feature representations $S(I^{t})$, $S(I^{t-1})$ and $S(I^{g})$, respectively. To represent the semantic correlation between the goal image and each of the input images in the 2D ego-centric view of the agent, we convolve the semantic visual representation of each input image with that of the goal image $\phi_{\hat{\theta}_{s}}(I,I_g) = S(I)\odot S(I^{g})$. Also, for each pair of consecutive input images $I_t$ and $I_{t-1}$, we compute the optical flow map $\psi(I_t,I_{t-1})$. We then stack the resulted pairs of collision maps $\phi_{\hat{\theta}_{c}}$, semantic correlation maps $\phi_{\hat{\theta}_{s}}$, and optical flow map $\psi$ to generate the visual representation $\Phi_{\hat{\theta}}$ at each timestep $t$. For brevity we omit $I$s from the equation.

\begin{equation}
\label{eq:rep}
\Phi_{\hat{\theta}_{t}} = [\phi_{\hat{\theta}_{c}, t},\phi_{\hat{\theta}_{c}, t-1},\phi_{\hat{\theta}_{s}, t},\phi_{\hat{\theta}_{s}, t-1},\psi_{t,t-1}] 
\end{equation}

At each timestep $t$, we feed the observation  $o_t = \Phi_{\hat{\theta}_{t}}$ into our policy learning module. Our policy learning module consists of a convolution layer of size $2 \times 2 \times 64$, and a ConvLSTM~\cite{xingjian2015convolutional} unit to incorporate the history. The policy is aimed to learn Q-values corresponding to the $k$ discrete actions. 

\subsection{Direct Policy Transfer to the Real World} 
 Our Semantic Net incorporates robust visual features trained on rich semantic object categories and is capable of producing correlation map between the visual goal and the robot observation. Our simple mechanism for computing the visual correlation between the goal and the observation is the crux of our network for modeling domain invariant stimuli $\phi_{\hat{\theta}_{s}}$. This stimuli is directly fed to our policy network so that the agent can learn generalizable policies that can be directly transferred to the real world. 

At the test time, we are able to use the same network for querying our visual servoing policy with semantic object categories. To do that, we mask the last layer of the Semantic Net with the category id $l$ to produce $S_l(I)$ and use it in lieu of the correlation map $\widehat{\phi_{\hat{\theta}_{s}}}(I,l) = S_l(I) $ which will be fed into our policy module. Given the fact that our Semantic Net is trained with real images and is robust to noisy samples~\cite{izadinia2018viser}, our approach for computing the semantic correlation map as an input to the policy network provides domain invariant representations which we will empirically show to work well in real world scenarios. In addition to that, our Collision Net is also domain invariant as it is pre-trained via domain randomization technique~\cite{sadeghi2017cadrl}.

\subsection{Object Reaching via Deep RL}
We consider a goal conditioned agent interacting with an environment in discrete timesteps. Starting from a random policy the agent is trained to choose actions towards getting closer to a goal. The goal is defined by cropping out the image patch around the goal object as seen at the initial state and is denote by $I^g$. At each timestep $t$, the agent receives an observation $o_t$, takes an action $a_t$ from a set of $k$ discrete actions $\{a^1,a^2,...,a^k\}$ and obtains a scalar reward $R_t$. Each action corresponds to a rotation angle using which a continuous motion vector is computed to move the agent forward in the 3D environment. The motion vector has constant velocity for all the actions. By following its policy $\pi_\theta$, the agent produces a sequence of state-action pairs $\tau=\{s_t,a_t\}^{T-1}_{t=0}$ after $T$ steps. The goal of the agent is to maximize the expected sum of discounted future rewards with a discounting factor $\gamma\in[0,1]$:

\begin{figure*}[t]
    \centering
    \includegraphics[width=\textwidth]{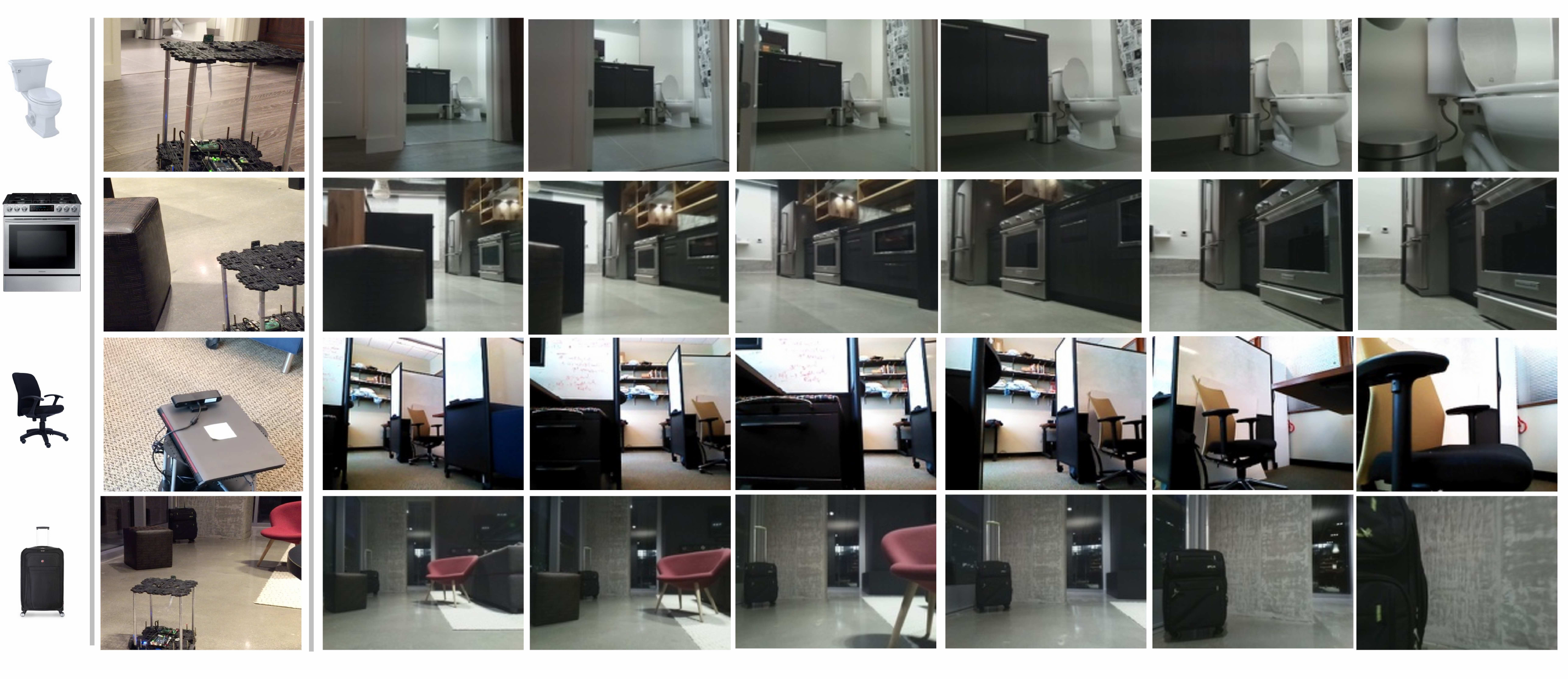}
     	\vspace{-0.35in}
    \caption{\small{Real world experiment for reaching semantic goals of toilet, teddy bear, chair and suitcase. The sequences show the robot view captured by head mounted monocular camera. DIViS can generalize to reach semantic goal objects in the real world.}}
    \label{real_exp_qual_sem}
     \vspace{-0.15in}
\end{figure*}

\vspace{-.14 in}
\begin{equation}
\label{eq:qfunc}
 Q(s_t,a) = R(s_t,a) + E_{\tau \sim \pi_\theta}[\sum_{t'=t+1}^{T}\gamma^{t'-t} R(s_{t'}, a_{t'})]
\end{equation}
\vspace{-.1 in}

\vspace{.07 in}
\noindent{\bf Dense Multi-Step Monte Carlo Rollouts:} 
We perform multi-step Monte Carlo policy evaluation~\cite{sutton1998reinforcement} for all possible $K$ actions at each state visited during an episode to generate dense rollouts. This enables us to train a deep network to make long-horizon dense predictions. Starting the agent from an initial state we generate dense rollouts with maximum length of $T$. For each state along the trajectory $\tau$, the dense rollouts are generated by performing $K-1$ additional rollouts corresponding to the actions $\{a^i\}_{i=1,a^i\neq a_t}^K$ which are not selected according to the agent's current policy $a_t \sim \pi_\theta$.

Figure~\ref{fig:net_arch}, demonstrates our dense Monte Carlo rollouts along a trajectory. The agent moves forward based on $\pi_\theta$. However, policy evaluations are computed for all possible actions that can be taken in each state. We evaluate the return of each action $a$ according to Equation~\ref{eq:qfunc}.
Therefore, for each state along the trajectories, we compute $\mathbb{Q}_{s_t} =\{Q(s_t,a^i)\}_{i=1}^{K}$  that densely encapsulates $Q$ values that quantify the expected sum of future rewards for each of the possible actions $a^i$. This policy evaluation provides us with a dataset of trajectories of the form:

\begin{equation}
\label{eq:trj}
(s_0,\mathbb{Q}_{s_0},a_0,...,s_{T-1},\mathbb{Q}_{s_{T-1}},a_{T-1})
\end{equation}

 If at any point during the episode or at any of the Monte Carlo branches, the agent collides with any of the objects in the scene the corresponding rollout branch will be terminated.

\vspace{.07 in}
\noindent{\bf Batch RL: } During training, we use batch RL~\cite{lange2012batch}, where we generate dense rollouts with Monte Carlo return estimates as explained above. Starting with a random policy, during each batch the agent explores the space by following its current policy to produce rollouts that will be used to update the policy. For each of the training tasks, we collect dense Monte Carlo rollouts multiple times each with a randomized rendering setup. During training, we collect samples from all our environments and learn a single policy over all different reaching tasks simultaneously. This enforces the agent to learn the common shared aspect between various tasks (i.e. to reach different goals) and acquire generalization capability to unseen test scenarios rather than memorizing a single task seen at the training time.

\vspace{.07 in}
\noindent{\bf Reactive Policy: } The \emph{reactive} agent starts from a random policy and does not save history from its past observations. The state at each timestep is described by the observation $s_t=o_t$. The visual observation $o_t$ at timestep $t$ is represented by $[\phi_{\hat{\theta}_{c}, t},\phi_{\hat{\theta}_{s}, t}]$ and $(o_t,\mathbb{Q}_{o_t})$ pairs are used to update the policy.

\vspace{.07 in}
\noindent{\bf Recurrent Policy: }
Starting from a random policy, the agent learns a recurrent policy using a sequence of observations. The recurrent policy uses the entire history of the observation, action pairs to describe the state $s_t = (o_1,a_1,...,a_{t-1},o_t)$. Each observation along the trajectory is represented by $\Phi_{\hat{\theta}_{t}}$ according to Equation~\ref{eq:rep}. Given the sequential nature of the problem, we use dense trajectories described in Equation~\ref{eq:trj} and we model the policy $\pi$ using a ConvLSTM unit. Intuitively, the history of past observations and actions will be captured in the hidden state of ConvLSTM.

\vspace{.07 in}
\noindent{\bf Reward Function: }
For the task of collision-free goal reaching, the agent should traverse a trajectory that decreases its distance to the goal object while avoiding obstacles. This implies two objectives that should be reflected in the reward function: (1) Collision avoidance: We define the collision reward function as $R_c = \min(1,\frac{d_o - r}{\tau_d - r}) $ to penalize the agent for colliding with various objects in the environment. Here, $r$ is the vehicle radius, $d_o$ denotes distance to the closest obstacle, and $\tau_d$ is a distance threshold. 
 (2) Progressing towards goal: The agent is rewarded whenever it makes progress towards the goal. Considering $d_t$ as the distance of the agent to the goal and $d_{init}$ as the initial distance of the agent to the goal, our progress reward function is defined as $R_g = \max(0,1-\frac{\min(d_{t},d_{init})}{d_{init}})$. The total reward is $R = R_c + R_g$.

\begin{figure*}[t]
    \centering
    \includegraphics[width=.99\textwidth]{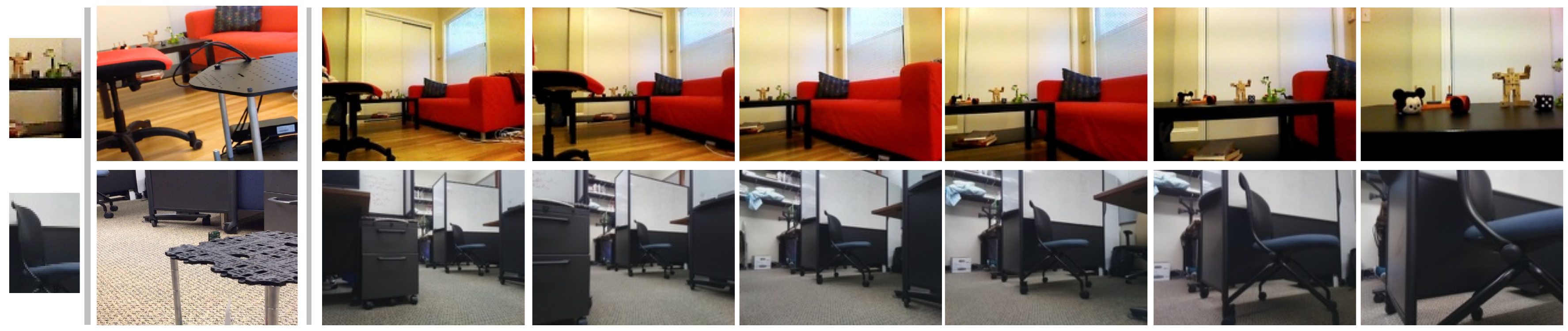}
     \vspace{-0.12in}
    \caption{\small{Real world experiments for reaching a goal in the robots view identified with a goal image. The first and second columns show the goal image and the third person view of the robot respectively. The image sequence show the input RGB images. Our goal reaching policy can generalize to diverse goals and can successfully avoid collisions in real world situations.}}
    \label{real_exp_qual_im}
         \vspace{-0.15in}
\end{figure*}

%% file: exp.tex
To evaluate the performance of DIViS for transferring semantic vision-based policy to the real world, we conduct real-world evaluation using two different real mobile robots with RGB cameras of different image quality. We test the capability of DIViS in reaching diverse goal objects in various unstructured real world environments while it avoids colliding with different types of furniture and obstacles. In addition, we study the various 
design decisions via detailed quantitative simulated experiments and compare against baselines, alternative approaches, and different network architectures.

\subsection{Real-World Evaluations}
\label{sec:realexp}
We use two different mobile robot platforms, TurtleBot2 and Waffle Pi (TurtleBot3), equipped with different monocular cameras (Astra and Raspberry Pi camera) to capture RGB sensory data used as input to our network shown in Figure~\ref{fig:net_arch}. 
We compare various settings of training DIViS  entirely in our domain randomized simulator and without any further fine-tuning or adaptation. Our goal is to answer the following key questions: (1) How well DIViS generalizes to real world settings while no real robot navigation data is used at training time? (2) How well does our approach transfer real world object-level semantics into the policy that is entirely trained in simulation? (3) How effective is our proposed recurrent policy compared to a reactive policy that is trained with similar pre-trained visual features? What is our performance compared to a conventional approach that visually matches the goal with current camera view? We study answer to these questions in the context of quantitative and qualitative real-world experiments.

\subsubsection{Quantitative Real World Experiments}
Our quantitative real world evaluation consists of two different setups for collision-free goal-reaching (a) Goal Image: reaching a visual goal as specified by a 
user selecting an image patch from the initial robot view. (b) Semantic object: Reaching a semantic object category such as \emph{chair}, \emph{teddy bear}, etc. that is inside the initial robot view.

\vspace{.06 in}
\noindent{\bf Generalizing to real world: } Table~\ref{tab:all_real_exp}, shows that our policy is robust to visually diverse inputs. Our Semantic Net and Collision Net and policy can directly transfer to real world  for reaching image goals while avoiding obstacles obtaining an average success rate of 82.35\% in goal image and 79.24\% in semantic object scenarios using two different real robot platforms.

\vspace{.06 in}
\noindent{\bf Generalizing to semantic objects:} Our experiments outlined in Table~\ref{tab:all_real_exp} and Table~\ref{tab:all_real_exp_cmpr} show that our policy can successfully generalize to reach semantic object categories with an average success rate of $79.24\%$ (over 53 trials) although it has not been directly trained for this task. Since our Semantic Net (explained in section~\ref{sec:approach}) is capable of localizing various object categories, it can provide the visual semantic correlation map for our policy network resulting in high generalization capability.

\vspace{.06 in}
\noindent{\bf Ablation and comparison with baseline:}
We compare the performance of DIViS to its reactive version explained in Section~\ref{sec:approach} as well as a Visual goal matching policy that mimics a conventional visual servoing baseline as explained in Section~\ref{sec:simexp_quant}.
To compare each method against DIViS, we run same scenarios with same goal and same initial robot pose. Each section in Table~\ref{tab:all_real_exp_cmpr}, compares the methods over similar scenarios. 
A successful policy should be capable of making turns to avoid obstacles while keeping track of the target object that can go out of the monocular view of the robot in sub-trajectories. Table~\ref{tab:all_real_exp_cmpr} outlines that DIViS has the highest success rate in all setups. Whilethe reactive policy can avoid obstacles it fails reaching the goal when it looses track of the objects at turns. Visual goal matching collides with obstacles more frequently as it greedily moves towards the object without considering the path clearance.  Having saved the memory of past observations via a recurrent policy, DIViS is able to keep track of the goal object when it gets out of the view and makes better decisions to avoid obstacles.

\begin{table}[t]
\begin{small}
\begin{center}
\small
\caption{\small{DIViS success rate using two real robot platforms. \label{tab:all_real_exp}} \vspace{-.05in}}
\renewcommand*{\arraystretch}{0.9}
\begin{tabular}[10pt]{lcccc}
\toprule
{Robot}& \multicolumn{2}{c}{Goal Image } & \multicolumn{2}{c}{Semantic Object} \\\cline{2-5}
&{success rate} & {\#trials} & {success rate} & {\#trials}  \\
\midrule
{WafflePi}   & {82.35\%} & {17} & {85.18\%} & {27}  \\
{TurtleBot2} & {81.81\%} & {22} & {73.07\%} & {26}  \\
\midrule
{Total}      & {82.35\%} & {39} & {79.24\%} & {53}  \\
\bottomrule
\vspace{-.4in}
\end{tabular}
\end{center}
\end{small}
\end{table}

\begin{table*}[t]
\begin{small}
\begin{center}
\small
\caption{\small{Comparing DIViS to baselines in reaching diverse goals in real world}\label{tab:all_real_exp_cmpr}\vspace{-.05in}}
\renewcommand*{\arraystretch}{0.9}
\begin{tabular}[10pt]{lcc|cc|c}
\toprule
& \multicolumn{2}{c}{Goal Image } & \multicolumn{2}{c}{Semantic object} & \multicolumn{1}{c}{Total} \\
&{success rate} & {\#trials} & {success rate} & {\#trials}  & {success rate}  \\
\midrule
{DIViS(ours)}   & {{\bf 83.78\%}} & {37} & {{\bf 75.6\%}} & {41}  & {{\bf 79.48\%}}\\
{DIViS-reactive(ours)} & {54.04\%} & {} & {43.9\%} & {}  & {48.71\%}\\
\midrule
{DIViS(ours)}   & {{\bf 82.35\%}} & {17} & {{\bf 85.18\%}} & {27}  & {{\bf 84.1\%}}\\
{Visual Goal Matching} & {35.29\%} & {} & {18.51\%} & {}  & {25.0\%}\\
\midrule
{DIViS(ours)}   & {{\bf 86.66\%}} & {15} & {{\bf 80.0\%}} & {15}  & {{\bf 83.33\%}}\\
{DIViS-reactive(ours)} & {53.0\%} & {} & {53.53\%} & {}  & {53.53\%}\\
{Visual Goal Matching} & {40.0\%} & {} & {20.0\%} & {}  & {30.0\%}\\
\bottomrule
\vspace{-.4in}
\end{tabular}
\end{center}
\end{small}
\end{table*}

\vspace{.05 in}
\noindent{\bf Comparison to ~\cite{pathak2018zero} for visual goal reaching:} Amongst prior navigation works, the most related paper to our work is~\cite{pathak2018zero} which servos to visual goals and is tested on a Turtlebot2 on 8 different scenarios in a single environment.~\cite{pathak2018zero} does not deal with simulation to real transfer and does not support navigating to semantic object categories . We tested our approach on 92 different scenarios conducted on 20 real world environments with substantially different appearance, furniture layouts and lighting conditions including outdoors. In total, we gained a success rate of 82.35\% in ``Goal image'' scenarios averaged over 39 trials.

\subsubsection{Qualitative Real World Experiments}
Figure~\ref{fig:teddy_expample} visualizes the performance of DIViS in a real wold ``semantic goal''  scenario where the goal is to reach the ``teddy bear''. This example demonstrates the importance of incorporating both object semantics and free space reasoning in choosing best actions to find a collision-free path in order to reach the goal object. First row in Figure~\ref{fig:teddy_expample} shows the RGB images observed by the robot and the second row shows the object localization score map for the ``teddy bear'' as obtained by our Semantic Net. Red arrows in the third row of Figure~\ref{fig:teddy_expample} show the action direction chosen by visual goal matching policy which only incorporates semantic object understanding. Green arrows show the action directions chosen by DIViS. While visual goal matching guides the robot to get close to the goal object, it does not have any mechanism to avoid obstacles and thus fails by colliding into other room furniture. On the other hand, DIViS maintains a collision-free path by choosing actions that both involve object semantics and free space reasoning. During traversal, DIViS may decide to take turns in order to avoid obstacles. This can result in loosing the sight of object for several steps. Being capable of maintaining a short memory, DIViS can turn back to the goal object after avoiding the obstacle. 

More qualitative examples of DIViS for \emph{goal image} and \emph{semantic object} such as ``toilet'', ``teddy bear'', ``chair'' and ``suitcase'' are provided in  Figure~\ref{real_exp_qual_im} and Figure~\ref{real_exp_qual_sem}, respectively. As demonstrated, DIViS can generalize to various real-world scenarios including diverse set of image goals, diverse object categories and and different environments with various indoor and outdoor layout and lighting. Please check supplementary videos at \href{https://fsadeghi.github.io/DIViS}{https://fsadeghi.github.io/DIViS} for more examples of the DIViS performance on two real robot platforms as well as results in our simulation environment.

\begin{figure*}[t]
\centering
\includegraphics[width=\linewidth]{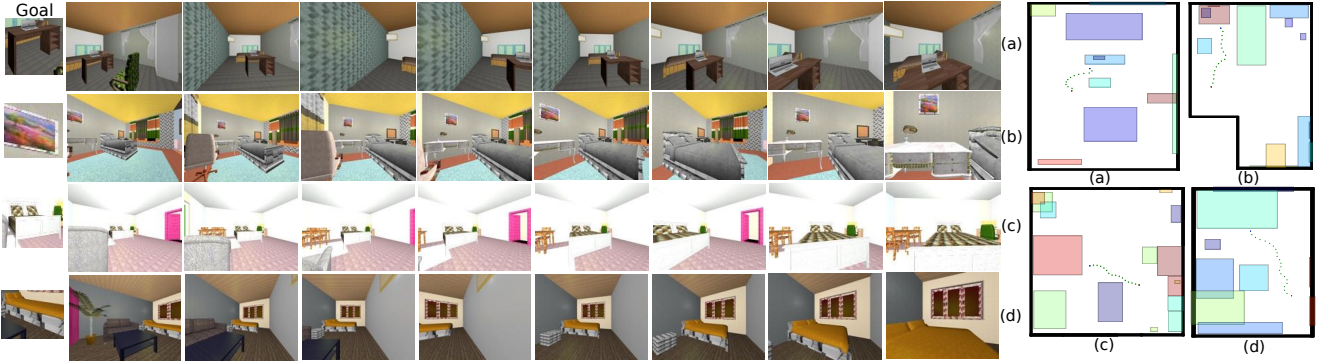}
 \vspace{-0.25in}
\caption{
   \label{fig:Qualitative results}
\small{Qualitative results on several complex test scenarios. In each scenario the trajectory taken by the agent is overlaid on the map (right) and several frames along the path are shown (left). To avoid collisions, the agent needs to take turns that often takes the goal object out of its view and traverse walkable areas for achieving the visually indicated goal in diverse scenarios.}}
\vspace{-0.15in}
\end{figure*}

\subsection{Simulation Evaluation}
\label{sec:simexp}
To generate simulation test scenarios, we sample free spaces in the environments uniformly at random to select the initial location and camera orientation of the agent. Therefore, the distance to the goal object and the initial view points change from one scenario to another. To further diversify the test scenarios and test the generalization capability, we do simulation randomization (also known as domain randomization)~\cite{sadeghi2017cadrl} in each test scenario using textures that were unseen during the training time. During the course of each trial, if distance of the agent to any of the scene objects other than the goal object becomes less than the agents radius~(i.e.~$\sim 16cm$) a collision event is registered and the trial is terminated.

\subsubsection{Quantitative Simulation Experiments}
\label{sec:simexp_quant}
For the evaluation criteria, we report \emph{success rate} which is the percentage of times that the agent successfully reaches the visually indicated object. If distance of the agent with the goal object is less than ~$30cm$, it is registered as success. We report the average success rate over a total of 700 different scenarios involving $65$ distinct goal objects collected from train( 380 scenarios) and novel test (320 scenario) environments . 
We compare DIViS (full model with recurrent policy and use of optical flow) against several alternative approaches explained below. Quantitative comparisons are summarized in Table~\ref{tab:sim_quant}.

\noindent{\bf Random Policy:} At each step, the agent selects one of the $k$ actions uniformly at random. 

\vspace{.07 in}
\noindent{\bf Visual Goal Matching:} This baseline models a greedy policy that follows an oracle rule of following the path with highest visual similarity to the goal and mimics conventional image-based visual servoing techniques to find the best matching visual features with the goal. Note that this policy uses a high-level prior knowledge about the underlying task while our agent s trained via RL does not have access to such knowledge and instead should learn a policy from scratch without any priors. Visual Goal Matching  selects one of the $k$ actions based on the spatial location of the maximum visual matching score of the visual goal and the current observation. To be fair, we use exact same Semantic Net pre-trained features used in our network to compute the correlation map.

\vspace{.07 in}
\noindent{\bf Visual Goal Matching with Collision Avoidance:}. This baseline combines prior sim2real collision avoidance method of~\cite{sadeghi2017cadrl} with conventional visual servoing for following the path with highest visual match to the goal and lowest chance of collision. We incorporate our predicted collision maps to extend Visual Goal Matching policy for better collision avoidance. Using our Collision Net and Semantic Net, we compute the spatial free space map and semantic correlation map for each observation. We sum up these two maps and obtain a single spatial score map that highlights the action directions with highest visual correlation and lowest chance of collision.
The agent selects one of the $k$ actions based on the spatial location with highest total score. To be fair, we use the exact same pre-trained features of Collision Net and Semantic Net in our network for this policy.  

\vspace{.07 in}
\noindent{\bf DIViS-Reactive policy:} The agent selects the best action based on the maximum Q-value produced by the reactive policy explained in Section~\ref{sec:approach}.

\vspace{.07 in}
\noindent{\bf DIViS-Recurrent:} The agent selects the best action based on the maximum Q-value produced by our recurrent policy without incorporating optical flow $\psi$ explained in Section~\ref{sec:approach}.
 
\vspace{.07 in}
\noindent{\bf DIViS-Recurrent with flow:} Our full model that select the best action based on the maximum Q-value produced by our network that also uses the optical flow between each two consecutive observations as explained in Section~\ref{sec:approach} and Equation~\ref{eq:rep}.
  
Table~\ref{tab:sim_quant}, compares the success rates between different approaches. The highest performance is obtained by our recurrent policy and the best results are obtained when optical flow$\psi$ is also incorporated. Interestingly, naive combination of collision prediction probabilities with visual goal matching results in lower performance than only using visual goal matching. This is because the collision avoidance tends to select actions that navigate the agent to spaces with smallest probability of collision. However, in order to reach goals the agent should be able to take narrow paths in confined spaces which might not have the lowest collision probability. Given the results obtained in this experiment, we used our recurrent policy with optical flow during our real-world experiments in Section~\ref{sec:realexp}.

\subsubsection{Qualitative Simulation Experiments}
We qualitatively evaluate the performance and behavior of the best policy i.e.  DIViS (with recurrent policy and optical flow) in a number of complex test scenarios. Figure~\ref{fig:Qualitative results} demonstrates several of such examples. 
In each scenario, the trajectory taken by the agent is overlaid on the top view of map for visualization. The initial and final position of the agent are shown by a red and a blue dot, respectively. During these scenarios, the agent needs to take actions which may increase its distance to the goal but will result in avoiding obstacles. However, the agent recovers by taking turns around the obstacles and successfully reaches the goal object.

\begin{table}[t]
\begin{small}
\begin{center}
\caption{\small Success rate in simulation. \label{tab:sim_quant}\vspace{-.05in}}
\begin{tabular}{l|cc}
\toprule
 {Method}  & {Seen Env.} & {Unseen Env.}\\
\midrule
{DIViS-Recurrent w/flow (ours)}      & \bf{87.6} & \bf{81.6}\\
{DIViS-Recurrent (ours)}              & 82.1 & 75.3\\
{DIViS-Reactive (ours)}               & 76.3 & 69.7\\
{Visual Goal Matching}             & 56.3 & 54.4\\
{Visual Goal Matching w/ collis.} & 48.9 & 47.8\\
{Random policy}          & 23.4 & 22.2\\
\bottomrule
\end{tabular}
\end{center}
\end{small}
\vspace{-.3in}
\end{table}

%% file: con.tex
In this paper, we described a novel sim-to-real learning approach for visual servoing which is invariant to the domain shift and is hence called domain invariant. Our proposed domain invariant visual servoing approach, called DIViS, is entirely trained in simulation for reaching visually indicated goals. Despite this fact, DIViS can successfully be deployed on real robot platforms and can flexibly be tasked to reach semantic object categories at the test time and in real environments. Our approach proposes transferring visual semantics from real static image datasets and learning physics in simulation. We evaluated the performance of our approach via detailed quantitative and qualitative evaluations both in the real world and in simulation. Our experimental evaluations demonstrate that DIViS can indeed accomplish reaching various visual goals and semantic objects at drastically different and unstructured real world environments. Our domain invariant visual servoing approach can lead into learning domain invariant policies for various vision-based control problems that involve semantic object categories. Future directions also include investigating how DIViS can be extended to work in dynamic environments such as ones with moving obstacles or goals which is an important challenge to be addressed in real-world robot learning.